\pdfoutput=1
\documentclass[11pt]{article}

\usepackage[final]{acl}

\usepackage{times}
\usepackage{latexsym}

\usepackage[T1]{fontenc}
\usepackage[utf8]{inputenc}

\usepackage{microtype}

\usepackage{inconsolata}

\usepackage{graphicx}

\usepackage{multirow}
\usepackage{amsmath}
\usepackage{booktabs}       % professional-quality tables
\usepackage{amsfonts}       % blackboard math symbols
\usepackage{nicefrac}       % compact symbols for 1/2, etc.
\usepackage{microtype}      % microtypography
\usepackage{xcolor}         % colors
\usepackage{marvosym}

\newcommand{\MODELNAME}{RwR}

\title{Recall with Reasoning: Chain-of-Thought Distillation for Mamba’s Long-Context Memory and Extrapolation}

\author{
  Jun-Yu Ma\thanks{Work done during intenship at Tencent AI Lab.}, Tianqing Fang\textsuperscript{\Letter}, Zhisong Zhang, Hongming Zhang, \\ \textbf{Haitao Mi, Dong Yu}\\ 
  Tencent AI Lab\\
  $^*$\texttt{junyu129@outlook.com}, \Letter \texttt{tianqfang@tencent.com}
  }
\begin{document}
\maketitle
\begin{abstract}
% For long-range sequence processing, as the context length increases, the Mamba model exhibits linear complexity compared to the quadratic complexity of the Transformer model, which effectively improves computational efficiency and saves computing resources.
% However, existing studies have shown that the performance of the Mamba model is significantly limited when processing long texts. In this paper, we propose the Distill Summary for Longer Memory (), which aims to use the chain of thought to improve the Mamba model's ability to find key information and thus improve long-term memory. We use the advanced LLaMa model to generate text-question-summary triples for training the Mamba model to improve the model's ability to extract information related to the question. In addition, in the inference stage, a multi-summary-then-answer strategy is designed to further improve the performance on long texts by limiting the length of the information processed each time. The experimental results on the LONGMEMEVAL dataset show that the proposed xx method effectively improves the model's long-term memory ability under the settings of 10k and 100k lengths.
Mamba's theoretical infinite-context potential is limited in practice when sequences far exceed training lengths. 
This work explores unlocking Mamba's long-context memory ability by a simple-yet-effective method, Recall with Reasoning (RwR), by distilling chain-of-thought (CoT) summarization from a teacher model. 
Specifically, RwR prepends these summarization as CoT prompts during fine-tuning, teaching Mamba to actively recall and reason over long contexts.
Experiments on LONGMEMEVAL and HELMET show RwR boosts Mamba's long-context performance against comparable Transformer/hybrid baselines under similar pretraining conditions, while preserving short-context capabilities, all without architectural changes.

\end{abstract}

\section{Introduction}
Transformer-based Large Language Models (LLMs)~\cite{DBLP:conf/nips/VaswaniSPUJGKP17,DBLP:journals/corr/abs-2302-13971,DBLP:journals/corr/abs-2307-09288} have demonstrated
significant capability in various real-world tasks, but suffer from quadratic complexity and poor length extrapolation.
% but they still face challenges in long-range language modeling due to quadratic computational demands and poor length extrapolation ability.
In contrast, Mamba~\cite{DBLP:journals/corr/abs-2312-00752} adopts a recurrent inference mode that ensures linear complexity and unlimited input length. 
However, despite having a theoretical capability of global memorization, 
empirical studies~\cite{DBLP:journals/corr/abs-2406-07887,ben2024decimamba,yelongmamba,DBLP:journals/corr/abs-2408-15496} have shown that Mamba struggles with long-context memory when the length of the processed text exceeds the model's training length.

To address this issue, efforts have focused on compressing unimportant tokens to reduce their negative impact.
DeciMamba \cite{ben2024decimamba} utilized the selective time-steps of Mamba to filter out unimportant tokens,
while ReMamba~\cite{DBLP:journals/corr/abs-2408-15496} used the similarity between query and tokens to select important tokens and remove unimportant ones.
However, in practice, long-context memory and extrapolation challenges persist when the input length significantly exceeds the training length of Mamba, even after applying such filtering.
Moreover, since the original input sentences are compromised, Mamba's language modeling performance may also be negatively impacted in this way (See Table~\ref{short-result}).

 \begin{figure}[t]
\centering
\includegraphics[width=0.44\textwidth]{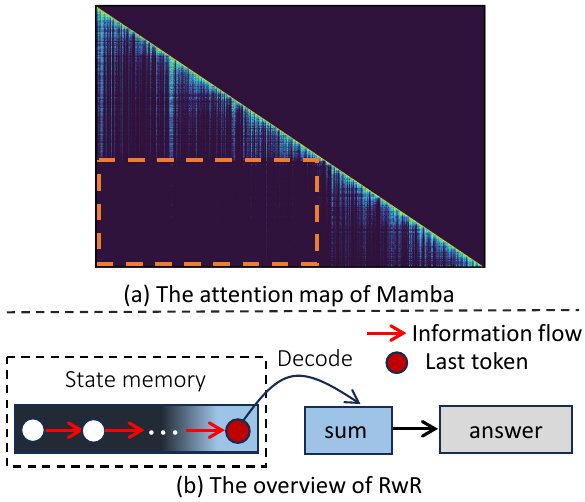}
\vspace{-2mm}
\caption{
(a) The simulated ``attention map'' of Mamba\protect\footnotemark. The orange rectangle shows that when Mamba encodes a long text, the representation of the current token is difficult to include the information of the previous token that is far away from it. (b) The state information of the text is gradually gathered from the beginning to the end to the last token.
RwR aims to first decode the last token state information to obtain a shorter summary, which can be fully accessed by Mamba when answering questions.}
\label{comparison}
\end{figure}

\footnotetext{A variant of self-attention in \citet{ben2024decimamba}.
}

% To address this issue, efforts have focused on removing or down-weighting unimportant tokens to reduce their negative impact.
% \citet{ben2024decimamba} proposed a non-training method to automatically detect and filter out unnecessary tokens, thereby reducing the length of the input.
% \citet{DBLP:journals/corr/abs-2408-15496} proposed a two-stage method, which compresses and selectively preserves critical information during an initial forward pass. \tq{Can you merge the previous two? They seem to use a discrete way of filtering out tokens, while Longmamba simply down-weight the unimportant ones.}
% \tq{\citet{Longmamba} XXX.}
% However, in practice, long-term memory issues still persist when the input lengths far exceeds the training length of Mamba even after such filtering.
% \tq{(Can you find proofs that these methods have a negative influence on short-context tasks?). Moreover, since the original input sentences are compromised, Mamba’s language modeling performance may also be negatively impacted. }

% Instead of changing the logic of state update, we propose to enhance the long-term memory ability of Mamba in a data-driven way. 
Instead of changing the logic of state update, we aims to unlock the long-context memory ability of Mamba in a data-driven chain-of-thought~\cite{wei2022cot, guo2025deepseek,DBLP:journals/corr/abs-2411-13504} paradigm.
Conceptually, our approach considers two key properties of Mamba: (1) its implicit attention mechanism naturally prioritizes recent tokens during decoding~\cite{ben2024decimamba} (Figure~\ref{comparison}(a)), and
(2) the state representation of the last token in the input theoretically encodes the complete history through selective state transitions (Figure~\ref{comparison}(b)).
Therefore,  
 Recall with Reasoning (\MODELNAME) is proposed, which teaches Mamba model to first decode relevant context from Mamba's fixed-size state memory, then performs reasoning based on this distilled historical summary.
Specifically, for a pair of long context and query, a more capable Transformer model is employed to generate relevant summary that are needed to answer the query. 
Such summary is then placed after the context and query as a new entry of training data for further fine-tuning.
% to generate context-question-summary triples for training the Mamba model, thereby enhancing its ability to extract question-related information in long texts.
% By augmenting a Supervised Fine-tuning (SFT) dataset with such query-aware summaries as CoT, model can ,the ability of recalling long-term memory of Mamba can be significantly improved.
By augmenting a Supervised Fine-tuning (SFT) dataset with such query-aware summary CoT, Mamba can effectively identify key information from long texts through CoT thinking, thus significantly improving its ability to recall long-context memory.
Furthermore, during the inference phase, a simple yet effective strategy of breaking down longer context into smaller pieces is used, which can further improve the performance.
% we propose a multi-summary-then-answer strategy to further optimize performance on extended texts by limiting the length of the information processed at each step. 

% To evaluate the long-term memory ability, we choose a chat-form benchmark LONGMEMEVAL~\cite{wu2024longmemeval}, and a comprehensive benchmark HELMET~\cite{DBLP:journals /corr/abs-2410-02694} covering seven application-centric categories.

To assess long-context memory and extrapolation ability, two benchmarks are adopted: a chat-form benchmark LONGMEMEVAL~\cite{wu2024longmemeval}, and an application-centric benchmark HELMET~\cite{DBLP:journals/corr/abs-2410-02694}.
Experimental results verify that {\MODELNAME} effectively improves the long-context memory ability of Mamba. 
% Here LONGMEMEVAL is challenging benchmark
% designed to evaluate the long-term memory ability of chat assistants, and HELMET is a comprehensive benchmark
% encompassing seven diverse, application-centric categories.
In addition, experiments on short-context language modeling tasks, like the RTE~\cite{DBLP:conf/mlcw/DaganGM05}, GSM8K~\cite{DBLP:journals/corr/abs-2110-14168}, Natural Question~\cite{DBLP:journals/tacl/KwiatkowskiPRCP19}, and SAMSum~\cite{gliwa-etal-2019-samsum}, show that our method does not affect the basic language modeling ability of Mamba.

% In summary, our contributions are as follows: (1) 

\section{Related Work}

\paragraph{State Space Models} 
% The structured State-Space Sequence (S4) model~\cite{DBLP:conf/iclr/GuGR22} is a novel
% alternative to transformers for modeling the long-term dependency.
% Based on S4, \citet{DBLP:journals/corr/abs-2312-00752} proposed a
% data-dependent SSM layer S6 and built a generic language model backbone Mamba~\cite{DBLP:journals/corr/abs-2312-00752}, which outperforms transformers of various sizes on real large-scale data and enjoys linear scaling in sequence length.
% Besides, \citet{DBLP:conf/icml/DaoG24} yields insights on how recent SSMs perform as well as Transformers on language modeling.
% Hybrid Mamba-Attention
% models like Jamba~\cite{DBLP:journals/corr/abs-2403-19887} and Samba~\cite{DBLP:journals/corr/abs-2406-07522} attempt to combine the
% benefits of attention mechanisms with Mamba’s efficiency in long-range modeling.

\citet{DBLP:conf/iclr/GuGR22} proposed the S4 model, which is a promising alternative to transformers for capturing long-term dependencies. 
Building on S4, \citet{DBLP:journals/corr/abs-2312-00752} introduced the data-dependent State Space Model (SSM) layer S6, and developed the Mamba~\cite{DBLP:journals/corr/abs-2312-00752} language model backbone. 
Mamba outperforms transformers of various sizes on large-scale real-world data and scales linearly with sequence length. 
Additionally, \citet{DBLP:conf/icml/DaoG24} provided insights into the performance of recent SSMs compared to transformers in the context of language modeling. 
Hybrid models like Jamba~\cite{DBLP:journals/corr/abs-2403-19887} and Samba~\cite{DBLP:journals/corr/abs-2406-07522} aim to combine the strengths of attention mechanisms with Mamba's efficient long-range dependency modeling.

% Additionally, some studies~\cite{DBLP:journals/corr/abs-2406-07887,ben2024decimamba,yelongmamba,DBLP:journals/corr/abs-2408-15496} have shown that Mamba models trained on length-limited contexts often
% experience performance degradation when extrapolated to longer contexts.
% {DBLP:conf/iclr/PengQFS24,DBLP:journals/corr/abs-2404-02060,ben2024decimamba}
 % have shown that the Mamba model struggles with long-term memory when the length of the processed text exceeds the model’s training length.

\paragraph{Long-context Memory} 
% Some studies~\cite{DBLP:conf/iclr/PengQFS24,DBLP:journals/corr/abs-2404-02060,ben2024decimamba} have shown that language models trained on length-limited contexts often experience performance degradation when extrapolated to longer contexts.
% Transformer-based models significantly encounter computational as context lengths increase.
% For Mamba models, DeciMamba~\cite{ben2024decimamba} proposed a non-training method to filter out some unimportant tokens, thereby reducing the length of the input.
% \citet{DBLP:journals/corr/abs-2408-15496} proposed to train a network to compress and selectively preserve critical information during an initial forward pass.
% \citet{yelongmamba} analyzes the root cause of the limitations of the Mamba models in handling long sequences and addresses this through a principled method applicable in various tasks.

Some studies~\cite{DBLP:conf/iclr/PengQFS24,DBLP:journals/corr/abs-2404-02060,ben2024decimamba} have highlighted that language models trained on fixed-length contexts tend to suffer performance degradation when applied to longer contexts. Transformer-based models, in particular, face significant computational challenges as the context length increases. 
For state space models, DeciMamba~\cite{ben2024decimamba} introduced a non-training method to filter out less important tokens, effectively reducing the input length. 
Additionally, \citet{DBLP:journals/corr/abs-2408-15496} proposed a technique where a network is trained to compress and selectively retain essential information during the initial forward pass. 
Meanwhile, \citet{yelongmamba} investigated the fundamental limitations of Mamba in handling long sequences and presented a principled approach to address these issues on various tasks.

 \paragraph{CoT Distillation} 
%  Migrating CoT capability into SLMs
% through distillation has attracted much attention.
% The approach typically employs CoT prompts to generate rationales from very large teacher models, and uses them to fine-tune small student models~\cite{DBLP:conf/acl/HoSY23,DBLP:journals/corr/abs-2311-06383}.
% \citet{DBLP:conf/acl/MagisterMAMS23} extensively explores the improvement of the reasoning ability of small models across multiple model architecture and observes
% the effects of student model size and data size on accuracy. 
% Apart from that, \citet{DBLP:conf/acl/WangWLGYR23}
% focus on generating more faithful and consistent
% rationales for CoT Distillation.

The distillation-based transfer of CoT ability to small language models (SLMs) has emerged as a prominent research direction. 
The predominant methodology leverages CoT prompting to extract rationales from large-scale teacher models (e.g., GPT-4), subsequently transferring these rationales to SLMs via fine-tuning~\cite{DBLP:conf/acl/HoSY23,DBLP:journals/corr/abs-2311-06383}. 
Building upon these foundations, \citet{DBLP:conf/acl/MagisterMAMS23} systematically investigated reasoning enhancement across multiple model architectures, empirically establishing the scaling laws governing student model ability and training data volume. 
In parallel, \citet{DBLP:conf/acl/WangWLGYR23} addressed the critical challenge of ensuring fidelity and consistency in rationale generation through constrained decoding optimization.

Compared with previous work~\cite{ben2024decimamba,DBLP:journals/corr/abs-2408-15496} that are the most relevant, a main difference should be highlighted.
They only focus on reducing sequence length by selectively removing
 unimportant tokens to improve the long-context memory ability of Mamba.
In contrast, our work makes the first attempt to use CoT distillation to directly improve the long-context memory ability of the Mamba without discarding the input tokens.
\section{Method}

\begin{figure}[t]
\centering
\includegraphics[width=0.48\textwidth]{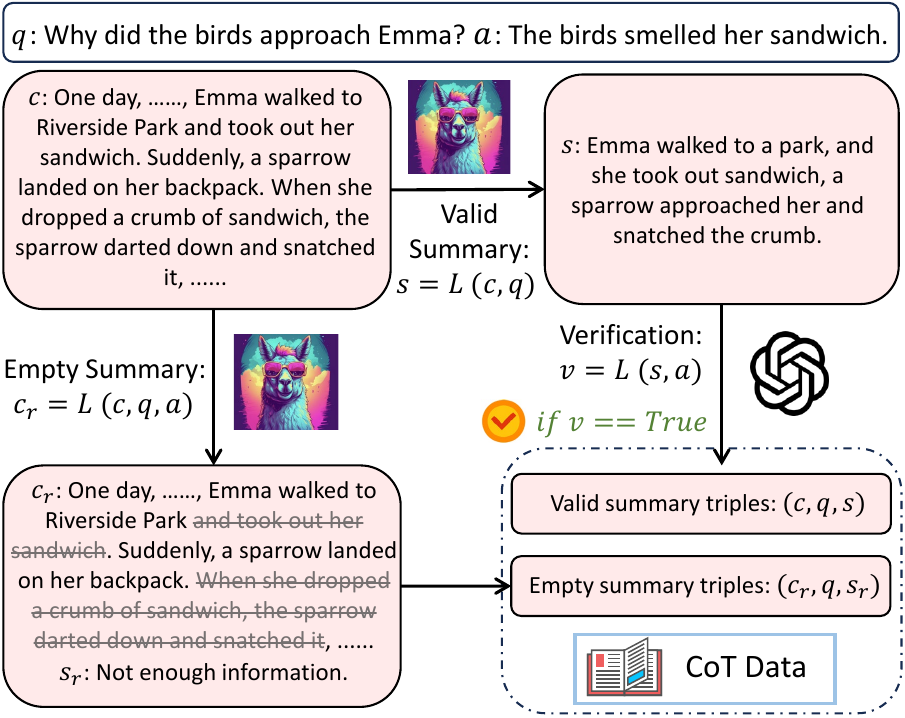}
\caption{
The data generation of {\MODELNAME}.
$c$ is the context, $q$ is the query for $c$, and $a$ is the ground-truth answer.
$c_r$ is the context in which the correct answer is removed.
$v$ refers to whether the valid summary contains correct answer.
$L$ refers to the LLM used in this step.
``......'' refers to a part of the context.
The gray part is the deleted text containing the answer to the query.
} 
\label{makedata}
\end{figure}

In this section, we present Recall with Reasoning (\MODELNAME).
The overview of the framework is presented in Figure~\ref{makedata}.
 % A teacher model is used to generate context-query-summary triples, which are subsequently adopted for training Mamba to enhance its extrapolation ability to
% extract query-related information in long texts.
% In addition, during the inference stage, a multi-summary-then-answer strategy is proposed to further improve the performance by fixing the input length processed at each step.

\subsection{Summary-based CoT Construction}

\paragraph{Valid Summary Extraction} Since Mamba~\cite{DBLP:journals/corr/abs-2312-00752}\footnote{https://huggingface.co/state-spaces/mamba-2.8b}  has not undergone instruction tuning, following \citet{DBLP:journals/corr/abs-2408-15496}, we utilized OpenOrca~\cite{DBLP:journals/corr/abs-2306-02707} as the SFT dataset, which primarily consists of question-answering data. 
Based on this dataset, context-query-summary triples are constructed to enhance the model’s ability to recall key information from long contexts.
 First, examples that contain background context relevant to the given query are selected from the dataset. 
Formally, let $c$ denote the context, $q$ the query, and $a$ the ground-truth answer. 
For simplicity, these selected examples are represented as $D = \{(c, q, a)\}$.
Next, as shown in Figure~\ref{makedata}, for each example $e \in D$, a Transformer model\footnote{In this paper, the Llama-3.1-8B-Instruct was used.} is employed to extract a summary of the context that is relevant to the query. 
The prompt used for this task is:
``<$c$> <$q$> Please extract a note relevant to the query:''
The extracted summary is denoted as $s$. 
To ensure the quality of the generated summary $s$, GPT-4o is used to verify their consistency with the correct answers $a$.
Samples with inconsistencies are filtered out.
Finally, the filtered examples are denoted as $D_f=\{(c,q,s)\}$.

\paragraph{Empty Summary Construction} 
% The summaries in the filtered triples often contain the correct answers, which can help the trained model extract useful information more effectively. 
In real-world scenarios, not all queries can be answered based on the context provided.
Training the model only on the above dataset may lead to overconfidence. 
For example, when context information is insufficient to answer the query, it will fail to answer: ``There is no enough information here'', but it will forcibly generate wrong answers.
% To mitigate this issue, we aim to enhance the model's ability to distinguish between relevant and irrelevant paragraphs.
% To achieve this, examples are constructed where the given context does not contain the answer to the query.
To mitigate this problem, we construct examples where the given context does not contain the answer to the query to enhance the model’s ability to distinguish between relevant and irrelevant paragraphs.
As shown in Figure~\ref{makedata}, $e \in D$ is selected, and then the Llama-3.1-8B-Instruct is used to locate the paragraphs in context $c$ that contain the correct answer.
These paragraphs are removed to obtain the modified context $c_r$.
This process results in a set of empty-summary data, denoted as $D_e = \{(c_r, q, empty)\}$.
Subsequently, the summary CoT data is the combination of $D_f$ and $D_e$, which is denoted as $D_s = [D_f, D_e]$.

Finally, Mamba is trained using the OpenOrca dataset along with the constructed dataset $D_s$.
In this way, the model can unlock the
ability of recalling from long input context via CoT.

% Moreover, we observe that the summaries in the filtered triples often contain correct answers, which can help the trained model extract useful information more effectively.
% However, in real-world scenarios, not all questions can be answered based solely on the provided context. Training the model directly on this dataset may lead to overconfidence, causing it to generate incorrect answers even when the available information is insufficient—ultimately resulting in hallucinations.
% To mitigate this issue, we aim to enhance the model’s ability to differentiate between relevant and irrelevant paragraphs. Specifically, we construct examples where the given context does not contain the answer to the question.
% To achieve this, we first select e∈De \in De∈D and use GPT-4o to identify the paragraphs in the context ttt that contain the correct answer. We then remove these paragraphs to obtain a modified context tdt_dtd​.
% This process results in a set of empty-summary data, denoted as De={(td,q,empty)}D_e = \{(t_d, q, \text{empty})\}De​={(td​,q,empty)}.
% Finally, we combine the filtered summary data DfD_fDf​ with the empty-summary data DeD_eDe​ to form the final dataset, denoted as Ds=[Df,De]D_s = [D_f, D_e]Ds​=[Df​,De​].

\subsection{Segmented Summarization for Answering} \label{strategy}
For scenarios with very long input lengths, a simple yet effective strategy is to segment longer context into smaller pieces.
First, the long context is divided into multiple parts.
Since the trained model has the ability to extract summaries, a summary is generated for each part.
Finally, these summaries are fed together into the model to answer the query.
This approach ensures that each processing step remains within a manageable length, which benefits the model’s memory. 
Moreover, Mamba has linear computational complexity as the input length increases, so this strategy does not increase the demand for computing resources.
\section{Experiments}

\begin{table}[t]
\footnotesize
\setlength{\tabcolsep}{1.2mm}
\begin{tabular}{lccccccc}
\toprule
\textbf{ Tasks }
% & \multicolumn{7}{c}{\textbf { ORCALE (10k)}} 
% & \multicolumn{7}{c}{\textbf { S (100k) }} 
% \cmidrule(r){2-8}
% \cmidrule(r){9-15}
 & \textbf{SU} & \textbf{SA} & \textbf{SP} &\textbf{MR} & \textbf{KU} & \textbf{TR} & \textbf{Avg}\\ 
\midrule
Orcale (10k) \\
\midrule
Mamba (Pre)  &0 &1.8  &0  &0.75  &0  &0  &0.4 \\
Mamba (SFT) &31.4 &39.2 &6.7 &8.3 &21.8 &14.3 &18.6 \\
DeciMamba &47.0 &41.1 &6.7 &3.0 &19.2 &14.3 &19.3 \\
ReMamba  &45.7 &33.9 &\textbf{13.3} &10.5 &28.2 &23.2 &24.4\\
\MODELNAME  &\textbf{48.6} &\textbf{44.6} &10.0 &\textbf{13.5} &\textbf{33.3} &\textbf{24.1} &\textbf{27.6} \\
\midrule
S (100k) \\
\midrule
Mamba (Pre)  &0 &0 &0 &0 &0 &0 &0 \\
Mamba (SFT) &11.4 &5.7 &0 &4.5 &10.3 &11.2 &8.0\\
\quad +SSA   &7.1  &7.1  &0  &5.3  &5.1  &9.8   &6.6\\
DeciMamba &0 &0 &0 &0 &0 &0.75 &0.2 \\
ReMamba   &2.9  &1.8  &0  &3.8  &7.7  &4.5  &4.0\\
\MODELNAME  &10.0  &7.1  &0  &6.3  &11.5  &\textbf{16.3}  &9.8  \\
\quad +SSA   &\textbf{12.9}  &\textbf{8.9}  &0  &\textbf{7.5}  &\textbf{16.7}  &15.0  &\textbf{11.4}  \\

% \midrule
% \textbf{MSD} &  \textbf{77.56} & \textbf{81.92} & \textbf{85.11} & \textbf{81.53} \\ 
% \midrule
% MSD w/o. distillers &  75.31  & 79.34  & 83.16  & 79.27  \\ 
% MSD w/o. $\mathcal{L}_{\text{MMD}}^{L}$ &  76.68  & 80.27  & 84.07  & 80.34  \\
% MSD w/o. $\mathcal{L}_{\text{MMD}}^{M}$ &  77.12  & 79.81  & 84.36  & 80.43  \\ 
% MSD w/o. all &  74.17  & 77.82  & 81.31  & 77.76  \\

\bottomrule

\end{tabular}
    \caption{The evaluation results (\%) on LONGMEMEVAL. The length of examples in ORCALE dataset is about 10k, and the length of examples in S dataset is about 100k. 
    ``SSA'' refers to the proposed strategy in Section~\ref{strategy}.
    Since the number of examples is different in different tasks, the Avg is the weighted average.}
    \label{longeval-result}
\end{table}

% \begin{table}[t]
% \footnotesize
% \setlength{\tabcolsep}{2.0mm}
% \begin{tabular}{lcccc}
% \toprule
%  \textbf{Datasets} & \textbf{Orcale (10k)} & \textbf{S (100K)} & \textbf{Avg} \\ 
% \midrule
% Mamba (Pre)  \\
% Mamba (SFT)  \\
% \quad +M \\
% ReMamba   \\
% xxx   \\
% \quad +M \\
% % \midrule
% % \textbf{MSD} &  \textbf{77.56} & \textbf{81.92} & \textbf{85.11} & \textbf{81.53} \\ 
% % \midrule
% % MSD w/o. distillers &  75.31  & 79.34  & 83.16  & 79.27  \\ 
% % MSD w/o. $\mathcal{L}_{\text{MMD}}^{L}$ &  76.68  & 80.27  & 84.07  & 80.34  \\
% % MSD w/o. $\mathcal{L}_{\text{MMD}}^{M}$ &  77.12  & 79.81  & 84.36  & 80.43  \\ 
% % MSD w/o. all &  74.17  & 77.82  & 81.31  & 77.76  \\

% \bottomrule

% \end{tabular}
%     \caption{xxx}
%     \label{conll-result}
% \end{table}

\begin{table}[t]
\centering
\footnotesize
\setlength{\tabcolsep}{3.0mm}
\begin{tabular}{lcccc}
\toprule
  \textbf{Tasks} & \textbf{RAG} & \textbf{ICL} &\textbf{SR} &\textbf{Avg} \\ 
\midrule
Mamba (Pre) &0 &0 &0 &0\\
Mamba (SFT)  &44.6 &39.0 &0.5 &28.0\\
DeciMamba  &39.8   &20.0   &\textbf{3.5} &21.1 \\
ReMamba  &41.8 &52.0 &1.7  &31.8 \\
\MODELNAME & \textbf{47.3}  & \textbf{54.0} & 1.6  & \textbf{34.3}  \\
% \midrule
% \textbf{MSD} &  \textbf{77.56} & \textbf{81.92} & \textbf{85.11} & \textbf{81.53} \\ 
% \midrule
% MSD w/o. distillers &  75.31  & 79.34  & 83.16  & 79.27  \\ 
% MSD w/o. $\mathcal{L}_{\text{MMD}}^{L}$ &  76.68  & 80.27  & 84.07  & 80.34  \\
% MSD w/o. $\mathcal{L}_{\text{MMD}}^{M}$ &  77.12  & 79.81  & 84.36  & 80.43  \\ 
% MSD w/o. all &  74.17  & 77.82  & 81.31  & 77.76  \\

\bottomrule

\end{tabular}
    \caption{The evaluation results (\%) on HELMET. The length of examples is set to 16k.}
    \label{helmet-result}
\end{table}

\begin{table}[t]
\footnotesize
\setlength{\tabcolsep}{1.1mm}
\begin{tabular}{lcccc}
\toprule
  \textbf{Tasks} & \textbf{Dialogue} & \textbf{NLI} &\textbf{Reasoning} &\textbf{Open-QA}  \\ 
\midrule
Mamba (Pre)  &25.2 &45.7 &71.5  &21.5 \\
Mamba (SFT) &23.3  &40.6 &88.5  &23.4  \\
DeciMamba & 24.7  &36.5 & 92.8 & 9.9 \\
ReMamba   &5.1  &\textbf{50.5}  &61.5   &21.4   \\
\MODELNAME &\textbf{28.1}  &46.9 &\textbf{93.0} &\textbf{23.7} \\
% \midrule
% \textbf{MSD} &  \textbf{77.56} & \textbf{81.92} & \textbf{85.11} & \textbf{81.53} \\ 
% \midrule
% MSD w/o. distillers &  75.31  & 79.34  & 83.16  & 79.27  \\ 
% MSD w/o. $\mathcal{L}_{\text{MMD}}^{L}$ &  76.68  & 80.27  & 84.07  & 80.34  \\
% MSD w/o. $\mathcal{L}_{\text{MMD}}^{M}$ &  77.12  & 79.81  & 84.36  & 80.43  \\ 
% MSD w/o. all &  74.17  & 77.82  & 81.31  & 77.76  \\

\bottomrule

\end{tabular}
    \caption{The results (\%) on short-context tasks.}
    \label{short-result}
\end{table}

\subsection{Experiment Settings}
\paragraph{Evaluation Datasets \& Metrics} For the evaluation of long-context memory and extrapolation, LONGMEMEVAL~\cite{wu2024longmemeval}, and an HELMET~\cite{DBLP:journals/corr/abs-2410-02694} were selected as benchmarks.
LONGMEMEVAL was a challenging benchmark
designed to evaluate the long-context memory ability of chat assistants across six memory tasks, including single-session-user (SU), single-session-assistant (SA), single-session-preference (SP), multi-session reasoning (MR), knowledge update (KU), temporal reasoning (TR).
% information extraction (IE), multi-session
% reasoning (MR), knowledge update (KU), temporal reasoning (TR), and abstaining on unanswerable questions (ABS).
It also contains two datasets of different length.
HELMET was a comprehensive benchmark for long-context language models covering seven diverse categories of tasks.
This paper selected several tasks in HELMET and set the evaluation length to 16k.
Besides, we also assessed the model's ability on short context setting, including four tasks and datasets: Dialogue (Mutual;~\citealp{DBLP:conf/acl/CuiWLZZ20}), NLI (RTE;~\citealp{DBLP:conf/mlcw/DaganGM05}), Reasoning (Natural Question;~\citealp{DBLP:journals/tacl/KwiatkowskiPRCP19}), and Open-domain QA ( Natural Question;~\citealp{DBLP:journals/tacl/KwiatkowskiPRCP19}).
Readers can refer to the Appendix~\ref{append-metric} for the details of evaluation metrics.

\paragraph{Training Details} Mamba-2.8b~\cite{DBLP:journals/corr/abs-2312-00752} was used as the backbone model.
OpenOrca~\cite{DBLP:journals/corr/abs-2306-02707} was used as instruction-tuning data.
To accommodate device memory constraints, the training examples were truncated to a maximum
length of 6,000 tokens.
Finally, 100,000 OpenOrca data and 10,000 constructed summary data were used for the proposed \MODELNAME.

\paragraph{Baselines} We included the untuned Mamba, Mamba fine-tuned 
 on the OpenOrca dataset, Decimamba~\cite{ben2024decimamba} and Remamba~\cite{DBLP:journals/corr/abs-2408-15496} as baselines.
Decimamba and Remamba were compression methods designed for the long-context memory of Mamba.
 
\subsection{Results}

\paragraph{Long-context Memory Tasks} Table~\ref{longeval-result} and~\ref{helmet-result} reported the long-context memory evaluation results on two benchmarks.
We observed that while previous long-context memory and extrapolation methods (e.g., DeciMamba and ReMamba) improve performance in the 10k length setting, their effectiveness decreased significantly in the 100k length setting and even underperformed directly fine-tuned Mamba (SFT).
 This suggests that existing methods have notable limitations in extending the Mamba’s context length.
In nearly all tasks, {\MODELNAME} enhanced the performance of Mamba across all context lengths.
This demonstrates that CoT can effectively extend the model’s processing length and improve its long-context memory ability.
Furthermore, for the 100k length settings, our SSA strategy further improved the performance for {\MODELNAME}. 
However, for the Mamba (SFT) that was trained without our constructed data, using this strategy results in a decrease in performance. 
This shows that our method can effectively improve the model's ability to extract summaries, thereby indirectly improving the model's long-context memory ability.

\paragraph{Short-context Tasks} 
In order to verify whether the proposed method has negative effects while improving long-context memory and extrapolation ability, several short-context tasks were selected for evaluation, and the results were shown in Table~\ref{short-result}. 
As shown in the table, compared with Mamba (SFT), the short-context language modeling ability of our method {\MODELNAME} has been slightly improved. 
However, the short-context abilities of DeciMamba and ReMamba were significantly reduced, which indicates that the previous compression methods affect the language modeling ability of Mamba and bring challenges in practical application.

\subsection{Other Architectures Study}
\paragraph{Extrapolation study} In order to verify the superiority of Mamba in extrapolation ability compared with other model architectures, length extrapolation experiments were conducted on the Transformer model and the hybrid SSM-Transformer model.
Specifically, we selected Transformer model Phi-2~\cite{javaheripi2023phi} and hybrid model Hymba~\cite{dong2024Hymba}, as they closely match Mamba in both the pre-training context length and model size.
Then these models were fine-tuned using the same data as those used by Mamba in this study. 
The performance of the fine-tuned Phi-2 and Hymba models was evaluated on the LONGMEMEVAL benchmark. 
As shown in Table~\ref{transformer-longeval-result}, at 10k length, the average performance of both Phi-2 and Hymba was slightly worse than that of Mamba. 
However, their performance was almost 0 at the 100k length setting, significantly lower than that of Mamba, which indicates that their length extrapolation ability is very limited.
In addition, the Phi-2 model only achieved certain performance on the single-session-user (SU)
and single-session-assistant (SA) tasks. 
These suggest that the Phi-2 model retains only some simple ability during extrapolating the length and is not generalizable.

\paragraph{Efficiency study} The rightmost column of Table~\ref{transformer-longeval-result} presents the average time taken by different models to process samples of varying dataset lengths. 
For a data length of 10k, the time differences between models were minimal, with the Transformer model requiring less time than the hybrid model. 
This is likely due to the more complex structure of the Hymba model, which requires additional processing steps. 
However, when processing data of length 100k setting, the Transformer model took significantly longer than the other models, which highlights the efficiency of the SSM model in processing long texts.

\begin{table}[t]
\footnotesize
\setlength{\tabcolsep}{0.8mm}
\begin{tabular}{lcccccccc}
\toprule
\textbf{ Tasks }
% & \multicolumn{7}{c}{\textbf { ORCALE (10k)}} 
% & \multicolumn{7}{c}{\textbf { S (100k) }} 
% \cmidrule(r){2-8}
% \cmidrule(r){9-15}
 & \textbf{SU} & \textbf{SA} & \textbf{SP} &\textbf{MR} & \textbf{KU} & \textbf{TR} & \textbf{Avg}   &\textbf{Time}\\ 
\midrule
Orcale (10k) \\
\midrule[0.5pt]
\MODELNAME  &48.6 &44.6 &10.0 &\textbf{13.5} &\textbf{33.3} &\textbf{24.1} &\textbf{27.6} &1.7s\\
Phi-2   &\textbf{61.4} &\textbf{67.8}  &0  &1.5  &5.1  &4.5 
&18.5 &2.5s\\
Hymba &40.0 &46.4 &13.3 &6.0 &35.9 &22.6 &24.8 &4.3s\\

\midrule
S (100k) \\
\midrule
\MODELNAME  &\textbf{10.0}  &\textbf{7.1}  &0  &\textbf{6.3}  &\textbf{11.5}  &\textbf{16.3}  &\textbf{9.8}  &10.8s \\
Phi-2   &0 &0 &0 &0.75 &1.3 &0 &0.4 &30.6s\\
Hymba &0 &0 &0 &0 &0 &0 &0 &17.5s\\

% \midrule
% \textbf{MSD} &  \textbf{77.56} & \textbf{81.92} & \textbf{85.11} & \textbf{81.53} \\ 
% \midrule
% MSD w/o. distillers &  75.31  & 79.34  & 83.16  & 79.27  \\ 
% MSD w/o. $\mathcal{L}_{\text{MMD}}^{L}$ &  76.68  & 80.27  & 84.07  & 80.34  \\
% MSD w/o. $\mathcal{L}_{\text{MMD}}^{M}$ &  77.12  & 79.81  & 84.36  & 80.43  \\ 
% MSD w/o. all &  74.17  & 77.82  & 81.31  & 77.76  \\

\bottomrule

\end{tabular}
    \caption{The evaluation results (\%) of SSM model \MODELNAME, Transformer model Phi-2 and hybrid model Hymba on the LONGMEMEVAL. ``Time'' refers to the average time consumed by each sample calculation.}
    \label{transformer-longeval-result}
\end{table}

% \begin{table}[t]
% \centering
% \footnotesize
% \setlength{\tabcolsep}{3.0mm}
% \begin{tabular}{lcccc}
% \toprule
%   \textbf{Model \& method} & \textbf{\MODELNAME} & \textbf{Hymba} &\textbf{Phi-2}  \\ 
% \midrule
%    \textbf{time} &10.6 &17.5 &30.8\\

% \bottomrule

% \end{tabular}
%     \caption{The evaluation results (\%) on HELMET. The length of examples is set to 16k.}
%     \label{helmet-result}
% \end{table}

\section{Conclusion}
This paper focuses on the long-context memory and extrapolation of Mamba.
A method called {\MODELNAME} is proposed to guide the CoT of Mamba to focus on summarizing and identifying key information in the previous context, thereby enhancing memory ability.
Experiments on the LONGMEMEVAL and HELMET datasets demonstrate that the proposed method effectively enhances the model’s long-context memory abilities, and in the meantime retaining the basic language modeling ability on other short-context tasks.
Further analysis shows that Mamba has better length extrapolation ability than the Transformer and hybrid models.
% \jy{extrapolation revise}

% \subsection{Appendices}

% Use \verb|\appendix| before any appendix section to switch the section numbering over to letters. See Appendix~\ref{sec:appendix} for an example.

\section*{Limitations}
There are several limitations for this paper.
First, this paper only conducts experiments on Mamba-2.8b, but whether it is effective on other SSM models such as Mamba2~\cite{DBLP:conf/icml/DaoG24} or Falcon mamba~\cite{zuo2024falconmambacompetitiveattentionfree} is still unknown and needs to be explored in future work.
Second, the longest test length in this paper is about 100k, but longer lengths, such as 200k, are not explored due to computational costs. 
Third, since the pre-training length of Mamba is limited to 2k, which is much shorter than more advanced Transformer models (such as Llama-3.3, which has a pre-training sequence length of up to 128k), the current Mamba model is not comparable to the state-of-the-art Transformer models.

% \section*{Acknowledgments}

\bibliography{custom}

\appendix
\section{Appendix}
\label{sec:appendix}

\subsection{Evaluation Metrics} \label{append-metric}
Following previous work, the LONGMEMEVAL~\cite{wu2024longmemeval}, 
was evaluated by GPT-4o. 
For the HELMET benchmark, the Retrieval-augmented generation (RAG) and  Synthetic recall (SR) tasks was evaluated by SubEM, and the Many-shot
 in-context learning (ICL) task was evaluated by Accuracy.
For short context tasks, the metrics were as follows:

  \textbf{Reasoning} on the GSM8K~\citep{DBLP:journals/corr/abs-2110-14168}, and the results were measured by solve rate.

  \textbf{Summarization} on the SAMSum~\citep{gliwa-etal-2019-samsum}, and the results were measured by the average of ROUGE-1, ROUGE-2 and ROUGE-L following.

  \textbf{Open-domain QA} on the Natural Question~\citep{DBLP:journals/tacl/KwiatkowskiPRCP19}, and the results were measured by exact match (EM) with the reference answer after minor normalization as in \citet{DBLP:conf/acl/ChenFWB17} and \citet{DBLP:conf/acl/LeeCT19}.

  \textbf{Natural language inference (NLI)} on the RTE~\citep{DBLP:conf/mlcw/DaganGM05}, and the results were measured by accuracy of two-way classification.

\end{document}